\documentclass[runningheads]{llncs}

\usepackage{eccv}

\usepackage{eccvabbrv}

\usepackage{graphicx}
\usepackage{booktabs}

\usepackage[accsupp]{axessibility}

\usepackage{hyperref}

\IfFileExists{orcidlink.sty}{\usepackage{orcidlink}}{\newcommand{\orcidlink}[1]{}}

\usepackage{amsmath}
\usepackage{xspace}
\usepackage{wrapfig}
\usepackage{enumitem}
\providecommand{\argmax}{}
\renewcommand{\argmax}{\mathop{\mathrm{arg\,max}}\limits}

\newcommand{\name}{{\textsc{SmartVL}}\xspace}

\newcommand{\squishlist}{
	\begin{list}{$\bullet$}
		{ \setlength{\itemsep}{0pt}
			\setlength{\parsep}{1pt}
			\setlength{\topsep}{1pt}
			\setlength{\partopsep}{0pt}
			\setlength{\leftmargin}{1em}
			\setlength{\labelwidth}{1em}
			\setlength{\labelsep}{0.5em} } }
\newcommand{\squishend}{\end{list} 
}

\usepackage[capitalize]{cleveref}
\crefname{section}{Sec.}{Secs.}
\Crefname{section}{Section}{Sections}
\Crefname{table}{Table}{Tables}
\crefname{table}{Tab.}{Tabs.}

\begin{document}

\title{Look Less, Think Faster: Joint Token-Compute Adaptation for Multimodal LLMs}

\titlerunning{Look Less, Think Faster}

\author{
Pengcheng Wang\inst{1}\orcidlink{0000-0002-2797-6973}
\and
Zhiquan Wang\inst{1}\orcidlink{0000-0002-5611-0335}
\and
Jayoung Lee\inst{1}\orcidlink{0000-0002-1011-6002}
\and
Zhuoyan Xu\inst{2}\orcidlink{0000-0001-5776-9388}
\and
Ran Xu\inst{3}\orcidlink{0000-0003-2913-9420}
\and
Saurabh Bagchi\inst{1}\orcidlink{0000-0002-4239-5632}
\and
Yin Li\inst{2}\orcidlink{0000-0003-4173-9453}
\and
Somali Chaterji\inst{1}\orcidlink{0000-0002-3651-6362}
}

\authorrunning{P.~Wang et al.}

\institute{
Purdue University, West Lafayette, IN, USA
\and
University of Wisconsin--Madison, Madison, WI, USA
\and
NVIDIA, Santa Clara, CA, USA
}

\maketitle

\begin{abstract}
Multimodal Large Language Models (MLLMs) have recently demonstrated strong performance across vision-language tasks. However, their high inference cost, arising from both the large number of input visual tokens and the heavy computation of the large language model (LLM), remains a key barrier to practical deployment. Recent work attempts to reduce the cost by adaptively optimizing individual dimensions, \eg, pruning redundant visual tokens or skipping LLM layers and heads. Nonetheless, prior approaches typically treat these dimensions independently and overlook a fundamental coupling: the available compute resources must be dynamically allocated across all dimensions based on the input content. To bridge the gap, we propose \textbf{\name}, a unified adaptive inference framework that jointly controls vision token number and model compute capability in response to varying input contents and compute budgets. \name introduces a vision-side token controller that dynamically selects informative visual tokens and an LLM-side compute controller that adaptively adjusts LLM computation. Importantly, these controllers are trained to coordinate with each other so that the overall inference cost satisfies a target budget. To allow this joint scheduling, we connect the controllers using a shared budget encoding and leverage a differentiable latency estimator for end-to-end training. This design enables \name to learn cross-stage allocation strategies that adapt to both input complexity and runtime compute constraints. Experiments across multiple MLLM benchmarks demonstrate that, with joint scheduling, \name consistently outperforms prior adaptive methods and achieves superior accuracy–efficiency Pareto frontiers. Project page: \href{https://www.schaterji.io/publications/2026/jointtokencompute}{https://www.schaterji.io/publications/2026/jointtokencompute}.

\end{abstract}

\section{Introduction}
\label{sec:intro}

\begin{sloppypar}
Multimodal Large Language Models (MLLMs) have become the dominant paradigm for multimodal understanding, achieving remarkable success across a wide range of vision-language tasks, including visual question answering (VQA), captioning, and grounding~\cite{liu2024improved, deitke2025molmo, tian2025drivevlm, zhang2024vision,Qwen-VL}. 
However, this generality comes at a high computational cost at inference time.
Importantly, this cost is largely fixed, regardless of the complexity of the input visual data. For example, processing an iconic image of an object requires essentially the same amount of computation as processing an image of the same object embedded in heavy background clutter. This mismatch becomes more pronounced under deployment scenarios with varying latency requirements. Real-time systems impose strict latency constraints\footnote{\small Throughout this work, we use a FLOPs-based compute budget as a hardware-independent proxy for these latency targets.} to ensure rapid responses, whereas offline systems may tolerate longer processing times. Even within a single deployment environment, the computational resources available to the model may fluctuate depending on system load.

Existing solutions for efficient inference typically adopt static inference pipelines~\cite{Qwen3-VL,deitke2025molmo}, which do not account for variability in input complexity or the compute budget available at runtime.
Consequently, these methods may allocate excessive computation to simple inputs, while lacking the flexibility to adapt when the available compute budget is significantly reduced at runtime, leading to suboptimal performance.
This raises a key question: \textit{can we design an adaptive inference framework for MLLMs that dynamically allocates computation based on both the input content and the available compute budget?}
\end{sloppypar}

To realize such an adaptive design, we must identify and control the primary drivers of MLLM inference cost. MLLMs often integrate a pretrained vision encoder with a large language model (LLM). The vision encoder transforms an input image or video into a sequence of vision tokens, which are further processed by the LLM alongside text tokens. This leads to a major computational bottleneck in the prefill stage, whose computation can be reduced across three primary dimensions. First, \textbf{sequence length}: a standard vision encoder such as CLIP ViT-L/14~\cite{radford2021clip} produces 576 patch tokens per image, all of which are concatenated with text tokens and processed through every layer of the language model. Second, \textbf{model depth}: the language model typically consists of 32 or more transformer layers, each contributing to latency and memory footprint. Third, \textbf{model width}: within each layer, computation is distributed across multiple attention heads and a wide feed-forward network. Maximizing performance under varying latency constraints and input content requires navigating the complex interplay between these computational axes.

\begin{sloppypar}
While prior work has exploited these individual axes through methods such as token pruning (\eg, FastV~\cite{chen2024fastv}, AIM~\cite{zhong2025aim}, LLaVA-PruMerge~\cite{shang2025llava}, VTW~\cite{lin2025vtw}) for sequence length, and adaptive layer/head skipping (\eg, AdaLLaVA~\cite{xu2025learning}, MoDES~\cite{huang2025modes}) for model depth and width, the primary limitation of these methods is that they address dimensions in strict isolation. This \textit{decoupled} approach leads to stage incompatibility and prevents globally optimal efficiency. Specifically, isolated token pruning evaluates spatial redundancy without awareness of the high-level linguistic context required by the LLM. Discarding visually redundant but semantically critical tokens causes irreversible information loss that propagates through downstream layers. Conversely, isolated model skipping assumes full-length input sequences, failing to adapt its computational depth when prior token pruning has already simplified the context. \textbf{Our key insight} is that token redundancy in the visual sequence and reasoning complexity within the LLM are \textit{fundamentally coupled}. The architectural capacity of the language model, specifically its depth and width, is strictly conditioned on the information density of the pruned visual input.
\end{sloppypar}

\begin{wrapfigure}[17]{r}{0.5\textwidth}
    \centering
    \includegraphics[width=1\linewidth, trim=0 6mm 0 15mm]{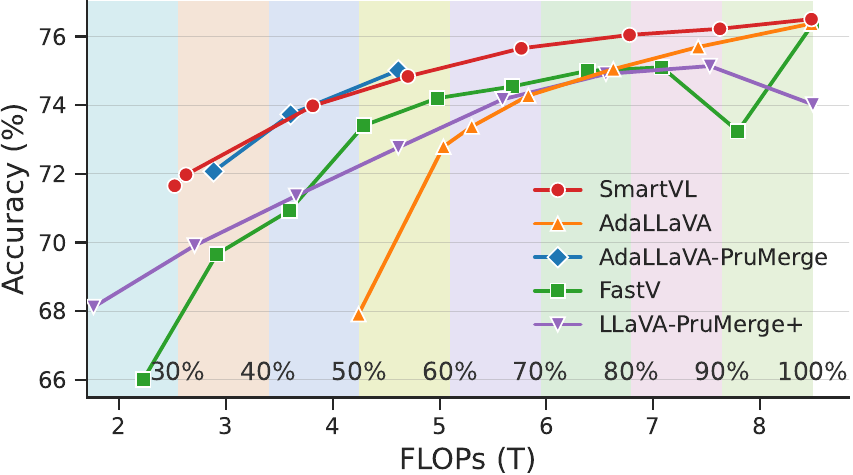}
    \caption{\small Pareto frontier on VQAv2. \name achieves a stronger accuracy--efficiency tradeoff than token-pruning methods (FastV and LLaVA-PruMerge+), compute-only control (AdaLLaVA), and the composed token+compute baseline (AdaLLaVA-PruMerge), highlighting the benefit of cross-dimensional adaptation.}
    \label{fig:vqav2_pareto}
\end{wrapfigure}

Driven by our insight, we propose \textbf{\name}, a unified adaptive inference framework that jointly controls sequence length, model depth, and model width. \name consists of two lightweight modules: a \textit{vision-side token controller} and an \textit{LLM-side compute controller}. Crucially, to ensure the framework is explicitly content- and budget-aware, these controllers formulate their allocation decisions based on fused intermediate features that jointly encode the input semantics and the target compute budget. Rather than relying on isolated heuristics, both controllers are trained end-to-end via a differentiable resource estimator that directly penalizes deviations from the compute budget.
By conditioning all computational decisions on the fused content and budget representations, the framework eliminates the need for manual threshold tuning. During inference, an integrated bounding mechanism strictly enforces the compute budget while enabling dynamic, cross-dimensional compute trade-offs. Consequently, the system autonomously decides for each input instance whether to allocate FLOPs toward retaining more visual context or activating deeper reasoning layers, as shown in Fig.~\ref{fig:vqav2_pareto}.

\begin{sloppypar}
We evaluate \name on multiple MLLM benchmarks (VQAv2, GQA, TextVQA, ScienceQA, POPE, MMBench, VizWiz) and compare against single-dimension token pruning baselines (LLaVA-PruMerge+, FastV), two-dimensional compute adaptive methods (AdaLLaVA), and a composed token-compute adaptation solution (AdaLLaVA-PruMerge).
Joint token-compute scheduling consistently yields superior accuracy versus efficiency Pareto fronts compared to individual token or compute controls or their naive combinations. Specifically, \name consistently outperforms AdaLLaVA across all FLOPs levels, achieving a 6.6\% accuracy gain at 50\% compute and maintaining this lead up to full capacity. While \name delivers comparable accuracy to PruMerge+ within its restricted 3T to 4.8T window, our framework scales across a significantly broader range from 2.5T to the full 8.5T capacity.
\end{sloppypar}

\smallskip
\noindent \textbf{Our main contributions} are summarized as follows.
\squishlist

\item We propose a unified adaptive inference framework that jointly controls sequence length, model depth, and model width to satisfy varying compute budgets. This approach explicitly addresses the fundamental cross-stage coupling between vision token spatial density and LLM architectural capacity that prior work overlooks.

\item We design two lightweight controllers linked by a shared budget encoding across the MLLM pipeline, jointly optimized via a differentiable latency estimator. To stabilize the compound discrete search space, we introduce a training strategy utilizing Gumbel-sigmoid sampling and an asymmetric budget violation loss, ensuring strict adherence to target latency while preserving reasoning accuracy.

\item Experiments across multiple MLLM benchmarks and diverse tasks demonstrate that our method consistently outperforms decoupled baseline solutions. Notably, our framework achieves a stronger Pareto frontier with a wider adaptation range: at the same computational budget, our method delivers higher accuracy. At $\sim$50\% compute budget, our method outperforms AdaLLaVA by an average of 7.8\% across seven benchmarks.
\squishend

\section{Related Work}
\label{sec:background}

\noindent\textbf{MLLMs architectures.}
Modern MLLMs adopt a ``patch tokens + LLM'' architecture: a ViT-style encoder generates dense visual tokens that are projected into the language model's embedding space and processed autoregressively~\cite{radford2021clip,jia2021scaling,liu2023visual}. Two primary approaches have emerged for the vision-to-language projection stage. The first employs learnable query mechanisms, such as the Q-Former in BLIP-2~\cite{li2023blip2} and InstructBLIP~\cite{dai2023instructblip}, that compress visual information through cross-attention with a fixed number of learnable queries. This query-based paradigm has been adopted in subsequent models such as VideoLLaMA~\cite{zhang2023videollama} and Qwen2-VL~\cite{Qwen2-VL}. Alternatively, LLaVA~\cite{liu2023visual} and MiniGPT-4~\cite{zhu2024minigpt} use a multilayer perceptron (MLP) to directly project visual tokens into the language model's embedding space, a design that has proven effective for multimodal instruction tuning and has been scaled up in recent models such as Molmo~\cite{deitke2025molmo} and Qwen2.5/3-VL~\cite{Qwen2.5-VL,Qwen3-VL}. Despite this architectural diversity, current MLLMs have a fixed computational cost regardless of the complexity of the input data, and cannot respond to varying compute budgets at inference time. These gaps motivate our content- and budget-aware adaptive inference.

\medskip
\noindent\textbf{Efficient inference of MLLMs.}
Compact architectures reduce inference cost by pairing small language backbones (\eg, Phi-2~\cite{javaheripi2023phi}) with vision encoders, as demonstrated by TinyGPT-V~\cite{yuan2023tinygpt}, LLaVA-$\phi$~\cite{zhu2024llava}, and TinyLLaVA~\cite{zhou2024tinyllava}. MobileVLM~\cite{chu2023mobilevlm} and MobileVLM V2~\cite{chu2024mobilevlmv2} further optimize backbone and projector design for mobile deployment, achieving competitive accuracy at sub-3B scales.
Post-training compression and conditional computation offer a complementary strategy. Quantization and low-rank adaptation can cut memory bandwidth and arithmetic intensity with minimal accuracy degradation~\cite{dettmers2023qlora,zhang2024quantvlm}. Mixture-of-experts (MoE) architectures such as MoE-LLaVA~\cite{lin2026moe} and LLaVA-MoD~\cite{shu2024llava} route each token to a small subset of expert modules, reducing total model capacity.
Another line of work targets the large number of visual tokens, which drive both attention complexity and key-value cache memory. FastV~\cite{chen2024fastv} and VTW~\cite{lin2025vtw} prune up to half of image tokens after shallow layers based on attention patterns. LLaVA-PruMerge~\cite{shang2025llava} performs early pruning before tokens enter the LLM using vision-encoder key-value statistics. IVTP~\cite{huang2024ivtp} combines vision-side and instruction-aware pruning to retain only query-relevant tokens. FastVLM~\cite{vasu2024fastvlm} replaces the standard ViT backbone with a hybrid encoder that natively produces far fewer tokens. While these methods largely improve efficiency, they inherit the fundamental static configuration problem and thus cannot support adaptive inference.

\medskip
\noindent\textbf{Adaptive, budget-aware inference.}
Several recent works seek to allocate computation dynamically: either in response to an explicit resource budget, or driven by the content of each input, or both.
MoDES~\cite{huang2025modes} targets adaptive computation in MoE-based MLLMs by learning per-token expert importance and applying input-dependent expert activation to reduce per-token computation. It focuses on expert-width control via lightweight gating without changing the base architecture. AIM~\cite{zhong2025aim} extends token reduction to a training-free setting, supporting token merging and pruning across a range of resource configurations within a single model. Most directly related to our work, AdaLLaVA~\cite{xu2025learning} learns an inference policy that, given an image-question pair and a latency budget proxy, adaptively drops transformer layers, attention heads, and MLP modules in the LLM while largely preserving accuracy. While AdaLLaVA demonstrates budget-conditioned model-component control, it does not jointly coordinate vision token count and model capacity. \name addresses this limitation with a unified framework that simultaneously controls all three axes of MLLM computation, conditioned on both input content and latency budget.

\section{\name: Adaptive Inference of MLLMs}
\label{sec:methods}

Given an image $I$ and a text prompt $T$, a standard MLLM encodes $I$ into $N$ visual tokens, concatenates them with text tokens, and processes the sequence through an $L$-layer transformer (hidden dimension $D$, $A$ heads per layer). The resulting prefill computation spans \textbf{three dimensions}: 1) sequence length $S$, determined by the retained visual tokens $S_{\mathrm{vis}}$; 2) model depth $L_{\mathrm{eff}}$, the number of executed transformer layers; and 3) model width $\alpha_l$, the active fraction of attention heads and FFN channels at layer $l$. We use $(S, L_{\mathrm{eff}}, \alpha_l)$ as the adaptive search space for our allocation framework.

\smallskip
\noindent\textbf{Budget constraint.}
Let $f(\cdot; \varphi)$ denote the MLLM with parameters $\varphi$. Let $b \in [b_{\min}, 1]$ denote a target compute budget expressed as a fraction of full-model prefill FLOPs. We use FLOPs as the budget metric because prefill computation is compute-bound and FLOPs provide a hardware-independent proxy for latency that correlates well with wall-clock time on modern accelerators. A minimum $b_{\min} > 0$ is enforced since the first $P$ layers (prefix layers, \cref{subsec:compute_controller}) always execute at full capacity.

\smallskip
\noindent\textbf{Adaptation policy.}
Given input $(I, T)$ and budget $b$, we seek a learned allocation policy $\pi$ that determines binary masks for three joint dimensions. These masks collectively define the computational configuration as follows:
\begin{itemize}[nosep]
    \item \textbf{Sequence length ($S$):} The token mask $\mathbf{m}_{\mathrm{token}}$ determines the number of retained vision tokens $S_{\mathrm{vis}} = \sum_i m_{\mathrm{token},i}$, yielding the effective sequence length $S = S_{\mathrm{text}} + S_{\mathrm{vis}}$, where $S_{\mathrm{text}}$ is the number of text tokens.
    \item \textbf{Effective depth ($L_{\mathrm{eff}}$):} The layer mask $\mathbf{m}_{\mathrm{layer}}$ dictates the execution of transformer blocks, directly controlling the vertical depth of the model.
    \item \textbf{Effective width ($\alpha_l$):} The activation fraction $\alpha_l \in [0,1]$ represents the per-layer width derived from $\mathbf{m}_{\mathrm{layer}}$ and $\mathbf{m}_{\mathrm{head}}$. Specifically, $\alpha_l = 0$ denotes a skipped layer, while $\alpha_l = 1$ indicates execution at full capacity.
\end{itemize}
At inference, the generated answer $X^a$ is produced by the model $f$ running under the specific allocation dictated by $\pi(I, T, b)$.
\begin{equation}
    X^a \leftarrow f\bigl(I, T;\; \varphi \,\big|\, \pi(I, T, b)\bigr),
    \label{eq:inference_under_pi}
\end{equation}
\ie, $f$ is applied to the prompt with token and layer/head selection given by $\pi$. The policy $\pi^*$ maximizes task accuracy subject to the budget constraint:
\begin{equation}
    \pi^* = \argmax\nolimits_{\pi} \; \mathcal{A}(I, T, f, \pi) \quad \text{s.t.} \quad \mathcal{C}(I, T, \pi) \leq b \cdot \mathcal{C}_{\mathrm{full}},
    \label{eq:budget_constrained}
\end{equation}
where $\mathcal{A}(I, T, f, \pi)$ denotes the accuracy of $f$ when run under allocation $\pi(I, T, b)$, $\mathcal{C}(I, T, \pi)$ denotes the resulting prefill compute cost, and $\mathcal{C}_{\mathrm{full}}$ is the full-model prefill cost. The policy $\pi$ is realized by two sequential controllers (\cref{subsec:token_controller,subsec:compute_controller}). Each transformer layer's cost scales as $\mathcal{O}(\alpha_l \cdot D^2 S + \alpha_l \cdot D S^2)$, where $D^2 S$ arises from linear projections (Q/K/V/O and FFN) and $D S^2$ from attention score computation. Aggregating over $L_{\mathrm{eff}}$ executed layers:
\begin{equation}
    \mathcal{C}_{\mathrm{prefill}} = \mathcal{O}\Bigl(\Sigma_{l=1}^{L}\ \alpha_l \cdot (D^2 S + D S^2)\Bigr).
    \label{eq:prefill_brief}
\end{equation}
Sequence length enters quadratically through attention, making token reduction the highest-leverage single dimension. Depth (number of active layers) and width ($\alpha_l$) each provide multiplicative reductions that compound with token reduction, so joint control over all three dimensions yields strictly larger savings than optimizing any single dimension alone.

\begin{figure}[t!]
    \centering
    \includegraphics[width=0.88\linewidth, trim=0 2mm 0 0]{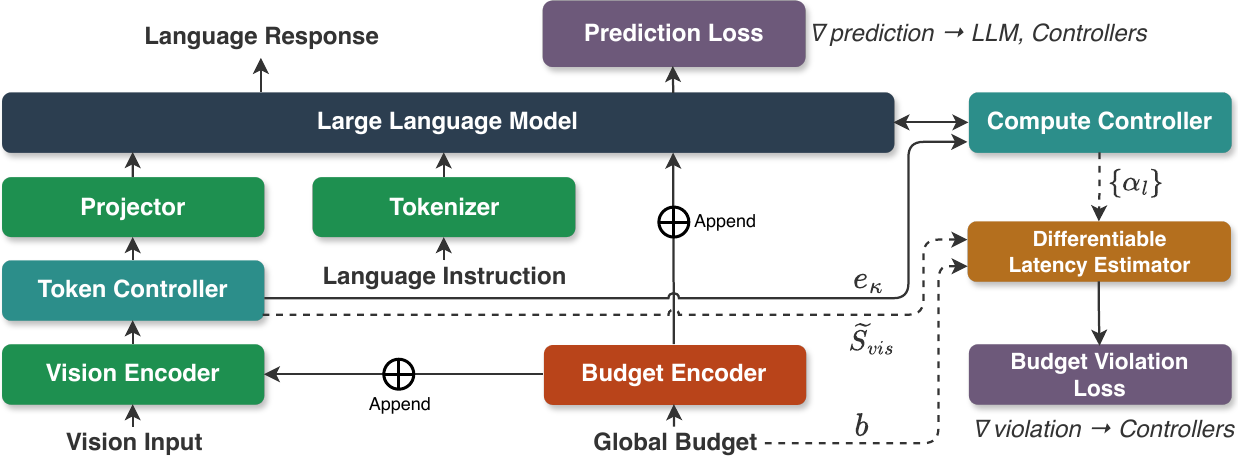}
    \caption{\textbf{Overview of \name.} A vision-side token controller and an LLM-side compute controller jointly govern three coupled compute dimensions: vision tokens (sequence length), active attention heads and FFN capacity (width), and executed transformer layers (depth). Both controllers are conditioned on the same global budget and trained end-to-end through a differentiable latency estimator, enabling them to learn content- and budget-aware allocation strategies across all three dimensions.}
    \label{fig:overview}
\end{figure}

\smallskip
\noindent\textbf{Framework overview.}
We realize this formulation with two learned controllers bridged by a shared budget signal (\cref{fig:overview}). A \textbf{vision-side token controller} (\cref{subsec:token_controller}) determines which visual tokens to retain via per-token Gumbel-sigmoid decisions. An \textbf{LLM-side compute controller} (\cref{subsec:compute_controller}) determines which transformer layers to execute and which attention head groups to activate within each layer. Both controllers are jointly optimized end-to-end through a \textbf{differentiable FLOPs estimator} (\cref{subsec:joint_optimization}) that maps their continuous outputs to an estimated compute ratio. Concretely, $\pi$ decomposes into two components:
\begin{equation}
    \mathbf{m}_{\mathrm{token}} = \pi_{\mathrm{tok}}(\mathbf{h}_b), \qquad
    (\mathbf{m}_{\mathrm{layer}},\, \mathbf{m}_{\mathrm{head}}) = \pi_{\mathrm{comp}}(\tilde{\mathbf{h}}_b^{(P)}),
    \label{eq:pi_decompose}
\end{equation}
where $\mathbf{h}_b \in \mathbb{R}^{C_v}$ (with $C_v$ representing the vision encoder hidden dimension) is the budget token's hidden state at the ViT output, encoding both image content and budget $b$ (\cref{subsec:token_controller}), and $\tilde{\mathbf{h}}_b^{(P)} \in \mathbb{R}^{D}$ is the budget token's hidden state after $P$ LLM prefix layers, fused with the token count $\kappa$ from $\pi_{\mathrm{tok}}$ (\cref{subsec:compute_controller}).

\subsection{Vision-Side Token Controller}
\label{subsec:token_controller}
As the initial stage of this joint allocation process, the token controller determines the sequence length by adaptively selecting a subset of vision tokens as the input to the LLM, conditioned on image content and compute budget.

\smallskip
\noindent\textbf{In-encoder budget fusion.}
To tightly couple token selection with visual processing and minimize overhead, we embed the budget signal directly into the vision encoder. The scalar budget $b$ is mapped to the ViT hidden space via sinusoidal encoding (PE)~\cite{vaswani2017attention} and an MLP:
\begin{equation}
\mathbf{e}_b^{\mathrm{ViT}} = \mathrm{MLP}_{\mathrm{ViT}}\bigl(\mathrm{PE}(b)\bigr) + \mathbf{p}_b 
\label{eq:budget_token_vit}
\end{equation}
where $\mathbf{p}_b$ is a learned position embedding.  This budget token is appended to the input sequence $\mathbf{X}_0 = [\mathbf{x}_{\mathrm{CLS}}, \mathbf{x}_1, \ldots, \mathbf{x}_N, \mathbf{e}_b^{\mathrm{ViT}}]$. By attending to all image patches throughout the ViT layers, it dynamically fuses visual content with the budget constraint to output a context-aware budget state $\mathbf{h}_b \in \mathbb{R}^{C_v}$.

\smallskip
\noindent\textbf{Per-token Gumbel-Sigmoid decisions.}
The token controller projects $\mathbf{h}_b$ via a linear layer to compute per-token logits $\mathbf{z} \in \mathbb{R}^N$. To enable differentiable binary routing, we apply the Gumbel-sigmoid relaxation~\cite{jang2017categorical, maddison2017concrete}. $z_i$ is perturbed by Gumbel noise $\xi_i$ to produce a continuous $y_i$ at temperature $\tau$:
\begin{equation}
y_i = \sigma\bigl((z_i + \xi_i) / \tau\bigr), \qquad
m_i = \mathbf{1}[y_i > 0.5] - \mathrm{sg}(y_i) + y_i.
\label{eq:gumbel_sigmoid}
\end{equation}
This straight-through estimator (STE), where $\mathrm{sg}(\cdot)$ is the stop-gradient operator, applies the discrete mask $\mathbf{m}_{\mathrm{token}} \in \{0,1\}^N$ during the forward pass while routing gradients through the continuous $y_i$ backward. The retained token count $S_{\mathrm{vis}} = \sum_i m_i$ thus adapts dynamically to the input. During training, dropped tokens are zeroed to maintain static tensor shapes for batching. At inference, we remove the noise ($\xi_i = 0$) and physically discard tokens where $\sigma(z_i) \leq 0.5$. The LLM processes only the selected tokens; RoPE~\cite{su2024roformer} natively handles this compressed, contiguous sequence without requiring positional re-encoding. Because the token controller is trained under the same budget signal used at inference time, its learned logits produce a token mask for each input from the image content and target budget, avoiding manually tuned retention ratios for each dataset.

\subsection{LLM-Side Compute Controller}
\label{subsec:compute_controller}
After receiving the visual sequence from the token controller, the LLM-side compute controller constitutes the second stage of the joint allocation framework.  It governs model depth $L_{\mathrm{eff}}$ and width $\alpha_l$ by dynamically deciding which transformer layers to execute and which attention head groups to activate within each layer. To explicitly address the cross-stage coupling, these discrete architectural decisions are strictly conditioned on the actual retained visual token count $S_{\mathrm{vis}}$ and the shared global budget.

\smallskip
\noindent\textbf{Budget encoding and prefix layers.}
To project the scalar budget $b$ into the LLM embedding space, we reuse the vision-side sinusoidal encoding but apply a distinct MLP to match the LLM hidden dimension $D$: $\mathbf{e}_b^{\mathrm{LLM}} = \mathrm{MLP}_{\mathrm{LLM}}(\mathrm{PE}(b)) \in \mathbb{R}^{D}$.  Appended to the end of the prompt sequence, this budget token receives standard RoPE and attends to all preceding visual and textual content. The initial $P$ transformer blocks (prefix layers) strictly execute at full capacity to establish a minimum computational baseline. Through these layers, the budget token aggregates global context to output a content-aware representation $\mathbf{h}_b^{(P)}$.

\smallskip
\noindent\textbf{Cross-stage token count embedding.}
Although the prefix state $\mathbf{h}_b^{(P)}$ captures visual context, it lacks an explicit measure of the consumed token budget. Because LLM inference cost strictly depends on the retained sequence length, we explicitly condition the compute controller on the normalized token survival rate $\kappa = \mathrm{sg}\bigl(\sum_{i=1}^N \sigma(z_i)\bigr) / N \in [0,1]$. This scalar is positionally encoded and added to the prefix boundary state:
\begin{equation}
\mathbf{e}_\kappa = \mathrm{MLP}_{\kappa}\bigl(\mathrm{PE}(\kappa)\bigr)
\qquad\tilde{\mathbf{h}}_b^{(P)} = \mathbf{h}_b^{(P)} + \mathbf{e}_\kappa.\label{eq:token_count_embed}
\end{equation}
Crucially, the stop-gradient operator $\mathrm{sg}(\cdot)$ isolates the vision and compute controllers' optimization trajectories. Backpropagating the LLM's architectural allocation gradients directly through $\kappa$ induces severe magnitude imbalances, prematurely forcing the vision controller into a degenerate sparse state. Detaching this specific path stabilizes the joint discrete optimization while rigorously preserving the forward budget conditioning.

\smallskip
\noindent\textbf{Gating via independent Gumbel-Sigmoid.}
Given the fused state $\tilde{\mathbf{h}}_b^{(P)}$, the compute controller predicts discrete execution masks for the remaining layers $P+1$ through $L$. 
Unlike prior adaptive methods (\eg, AdaLLaVA) that enforce a deterministic top-$K$ constraint via Gumbel-softmax selection, our framework applies independent Gumbel-sigmoid activations (\cref{eq:gumbel_sigmoid}) simultaneously to each layer or head group. This independent formulation is essential because the optimal active module count is not statically predefined; rather, it emerges dynamically from the global budget and the preceding vision token survival rate. To prevent premature structural collapse, we initialize the linear gating biases positively, ensuring the network begins optimization near full capacity before the asymmetric violation loss (\cref{subsec:joint_optimization}) induces selective deactivation. We explore two specific allocation granularities:
\noindent{\textbf{Layer controller (L)}.}
A single linear projection maps the fused state to $L-P$ logits. These are processed via the independent Gumbel-sigmoid function to produce a binary layer mask $\mathbf{m}_{\mathrm{layer}} = (g_l)_{l=P+1}^L \in \{0,1\}^{L-P}$. Each element $g_l$ governs the execution of its corresponding dynamically scheduled layer $l$: if $g_l = 0$, the layer is strictly bypassed via its residual connection; if $g_l = 1$, it executes at full capacity ($\alpha_l = 1$). 

\smallskip
\noindent{\textbf{Layer-Head controller (LH)}.}
For finer-grained allocation, the $A$ attention heads are partitioned into $G$ equal-sized groups (\eg, $G{=}4$ for LLaMA-7B with $A{=}32$). The linear gating head generates a matrix of group-level binary masks for the dynamically scheduled layers with $\mathbf{M} = \mathrm{GumbelSigmoid}\bigl(\mathbf{W}_h\, \tilde{\mathbf{h}}_b^{(P)} + \mathbf{c}_h\bigr)$.
Each row in $\mathbf{M}\in \{0,1\}^{(L-P) \times G}$ corresponds to a layer $l$, denoted as $\mathbf{M}_l \in \{0,1\}^G$, and serves as a binary mask broadcast to its respective head groups within that layer; thus $\mathbf{M} = \mathbf{m}_{\mathrm{head}}$. To maintain balanced computational reduction, one decision per group synchronously controls both the multi-head attention components and the corresponding intermediate dimension of the FFN. The active capacity fraction for layer $l$ is dynamically computed as $\alpha_l = \sum_{j=1}^{G} M_{l,j}/G$. Consequently, when all group masks for a specific layer evaluate to zero ($\alpha_l = 0$), the entire layer is skipped, effectively unifying depth and width control within a single formulation.

\subsection{Joint Training and Inference}
\label{subsec:joint_optimization}
Jointly optimizing the allocation decisions across both the vision token controller and LLM compute controller under a strict global budget poses fundamental challenges regarding non-differentiability and training stability. 
To circumvent the discrete bottleneck and map surrogate probabilities to a continuous compute ratio, we design a differentiable FLOPs estimator. Guided by this estimated ratio, an asymmetric budget violation loss is applied to ensure stable, soft-guided optimization without inducing structural collapse. Furthermore, a dynamic budget sampling procedure is integrated during training, enabling the unified model to generalize across a continuous spectrum of resource constraints.

\smallskip
\noindent\textbf{Differentiable latency estimator.}
With $\tilde{S}_{\mathrm{vis}} = \sum_{i=1}^{N} \sigma(z_i)$ (the differentiable proxy for $S_{\mathrm{vis}}$), $\alpha_l$ the effective layer activity, and $S = S_{\mathrm{text}} + \tilde{S}_{\mathrm{vis}}$, the estimated prefill cost is calculated as:
\begin{equation}
\hat{\mathcal{C}} = \Sigma_{l=1}^{L}\ \alpha_l \bigl[(8D^2 + 6DD_{\mathrm{int}}) \cdot S + 4D \cdot S^2\bigr] + 2DV \cdot S,
\label{eq:compute_est}
\end{equation}
where $D_{\mathrm{int}}$ is the FFN intermediate dimension and $V$ the vocabulary size. The normalized ratio $r = \hat{\mathcal{C}} / \mathcal{C}_{\mathrm{full}}$ is differentiable in both $\tilde{S}_{\mathrm{vis}}$ and $\{\alpha_l\}$. Let $\theta$ denote the parameters of both controllers, including their respective budget and token-count embedding modules. This differentiable latency estimator bypasses the non-differentiable discrete routing bottleneck, establishing an end-to-end gradient path for the budget constraint. The gradient of the budget penalty ($\partial r / \partial \theta$) propagates exclusively to these controller parameters, explicitly driving the vision token controller and LLM compute controller to learn budget compliance. 

\smallskip
\noindent\textbf{Asymmetric budget violation loss.} Given the estimated compute ratio $r$ and target budget $b$, we formulate an asymmetric penalty:
\begin{equation}
\mathcal{L}_{\mathrm{over}} = \max(0, r - b)^2, \quad 
\mathcal{L}_{\mathrm{under}} = \max(0, (b - \mu) - r), \label{eq:loss_over_under}
\end{equation}
where the margin $\mu$ defines a zero-penalty tolerance interval $[b{-}\mu,\; b]$. The asymmetric form reflects the different consequences of the two violations. Overshooting the budget breaks the requested operating point and is therefore penalized quadratically, while under-utilization remains feasible but leaves available capacity unused. We penalize the latter linearly, providing a steady incentive to exploit compute without overwhelming the task prediction loss. The final training objective integrates these terms:
\begin{equation}
\mathcal{L}_{\mathrm{total}} = \mathcal{L}_{\mathrm{pred}} + \lambda(t) \cdot (w_{\mathrm{over}} \cdot \mathcal{L}_{\mathrm{over}} + w_{\mathrm{under}} \cdot \mathcal{L}_{\mathrm{under}}),\label{eq:total_loss}
\end{equation}
where $w_{\mathrm{over}}$ and $w_{\mathrm{under}}$ are weights for $\mathcal{L}_{\mathrm{over}}$ and $\mathcal{L}_{\mathrm{under}}$, respectively. $\lambda(t) = \min(1,\; t / T_{\mathrm{warmup}})$ linearly scales the violation weight. This scheduled warmup is critical for optimization stability. Enforcing an immediate budget constraint before the controllers learn to distinguish task-relevant features causes the penalty gradients to dominate $\mathcal{L}_{\mathrm{pred}}$, prematurely forcing the network into degenerate sparse states. Delaying full regularization ensures the controllers first establish reliable importance distributions, grounding subsequent pruning in learned task utility rather than random initialization.

\smallskip
\noindent\textbf{Training procedure.} At each step, a target budget $b$ is sampled uniformly from $[b_{\min}, b_{\max}]$ and applied to the entire batch. This enables generalization across compute budgets, while batch-level sharing maintains hardware execution efficiency. To achieve fine-grained sensitivity to $b$, we employ sinusoidal encodings that allow the controllers to handle unseen budget ratios. The framework undergoes end-to-end optimization of the LLM, projector, and both controllers to coordinate decisions across the perception and cognition stages; the vision encoder remains frozen to ensure a stable input feature distribution for training. 

\smallskip
\noindent\textbf{Inference.}
Removing Gumbel noise yields deterministic execution masks at inference. Because the differentiable training objective enforces the budget softly, the raw masks can occasionally exceed the target budget; for $b<1.0$, we therefore apply a deterministic budget projection. Empirically, we find that generative quality is relatively resilient to sequence truncation, so the projection first discards the lowest-confidence visual tokens. It reduces layers or heads only when token-level adjustment alone is insufficient.

\smallskip
\noindent\textbf{Implementation details.} Controller hyperparameters, including the token and compute Gumbel temperatures, are fixed across benchmarks and budgets; each operating point is selected by changing the target compute budget at inference.

\section{Experiments and Results}
\label{sec:evaluation}

\subsection{Experiment Setup}
\label{subsec:exp_setups}

\noindent\textbf{Benchmarks.}
We evaluate zero-shot performance via \texttt{lmms-eval}~\cite{zhang2025lmms} on seven benchmarks across diverse capabilities: VQAv2~\cite{goyal2017making} (general), GQA~\cite{hudson2019gqa} (spatial), TextVQA~\cite{singh2019towards} (text-centric), ScienceQA~\cite{lu2022learn} (reasoning), VizWiz~\cite{gurari2018vizwiz} (real-world blind), POPE~\cite{li2023evaluating} (hallucination), and MMBench~\cite{liu2024mmbench} (comprehensive). Following prior works~\cite{xu2025learning}, we report official metrics and average prefill FLOPs.

\smallskip
\noindent\textbf{Baselines.}
We compare \name against four methodological categories: (1) \emph{Compute-only} (AdaLLaVA~\cite{xu2025learning}) featuring LLM-side adaptation only; (2) \emph{Token-only} (LLaVA-PruMerge+~\cite{shang2025llava}, FastV~\cite{chen2024fastv}) focused on sequence pruning with a static LLM; (3) \emph{Sequential} (AdaLLaVA-PruMerge), a two-stage composition lacking joint end-to-end optimization; and (4) \emph{Full-model reference} (LLaVA-1.5~\cite{liu2024improved}) as the performance upper bound and FLOPs normalization anchor.
For fixed token-pruning baselines, we use their original interfaces to sweep preset retention ratios up to full token retention, covering 10\%--100\% for FastV and 1/8--1 for PruMerge+. The resulting fixed-ratio configurations form the operating points used for their Pareto envelopes.

\smallskip
\noindent\textbf{Model variants.}
We evaluate two variants of our design, differing in the compute controller granularity: \name-L uses layer-level gating (depth only) and \name-LH uses head-group gating (depth and width). Both variants include the vision-side token controller.

\subsection{Main Results and Discussion}
\label{subsec:main_results}

\noindent\textbf{Accuracy-efficiency tradeoffs.}
We previewed the results on VQAv2 in \cref{fig:vqav2_pareto}; here we provide quantitative comparisons. On VQAv2, \name shows consistent gains across budgets. At 50\% FLOPs, \name reaches 74.4\% accuracy, compared with 67.9\% for AdaLLaVA and 74.5\% for AdaLLaVA-PruMerge. AdaLLaVA suffers a sharp drop at 50\% budget (its minimum operating point), whereas \name maintains strong accuracy from 20\% to 100\% of full budget. At 70\% FLOPs, the gap between \name (75.7\%) and AdaLLaVA (74.4\%) remains 1.3 points, while at full budget \name matches the base model (76.5\%).
AdaLLaVA-PruMerge achieves comparable accuracy in the mid-range but is limited to a few manually configured token-retention ratios (we test 25\% PruMerge+ token retention with AdaLLaVA layers at 65\%, 80\%, and 100\%) and cannot adapt continuously. FastV and LLaVA-PruMerge+ are fixed token-pruning methods (we sweep their original retention-ratio interfaces as described in \cref{subsec:exp_setups}); they use a dataset-level retention configuration and cannot adaptively adjust for each request or sample. These results confirm that joint cross-dimensional control yields a better Pareto frontier than decoupled baselines.

\begin{figure}[t!]
    \centering
    \includegraphics[width=1\linewidth, trim=0 6mm 0 0]{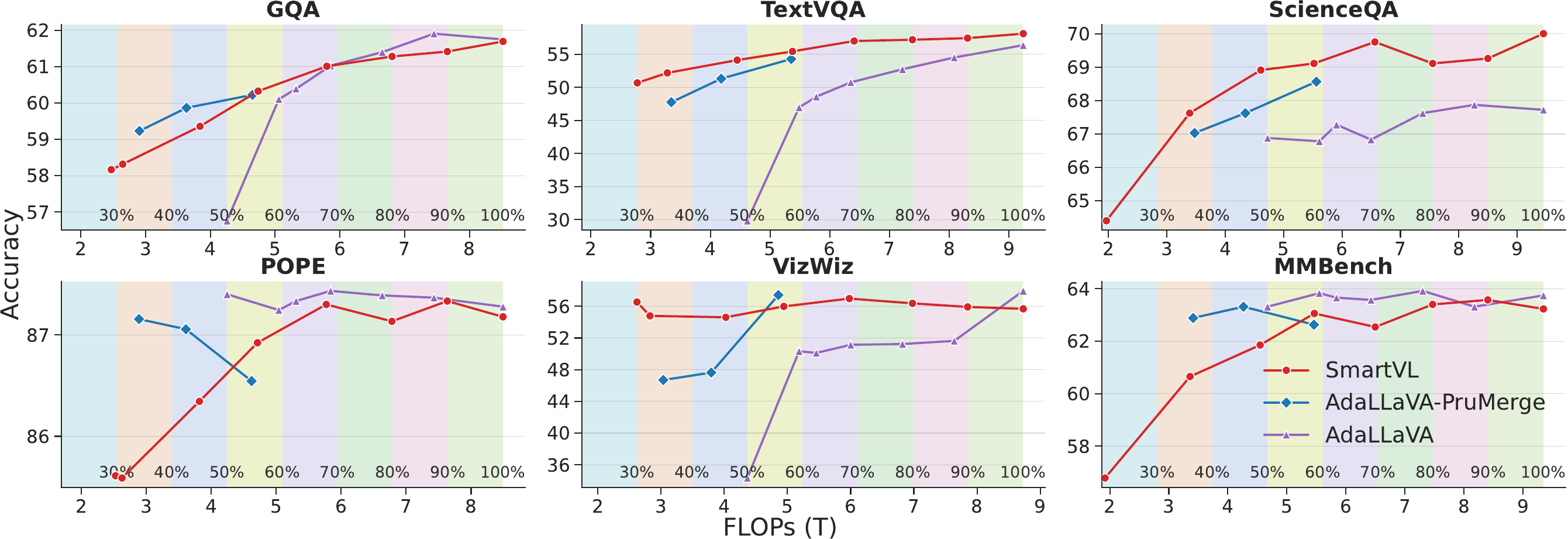}
    \caption{\textbf{Accuracy--Latency Pareto comparison across benchmarks.} Pareto fronts on (a) GQA, (b) TextVQA, (c) ScienceQA, (d) POPE, (e) VizWiz, and (f) MMBench. \name consistently achieves a stronger or comparable accuracy--efficiency tradeoff across operating points against AdaLLaVA and AdaLLaVA-PruMerge.}
    \label{fig:pareto_main}
\end{figure}

\smallskip
\noindent\textbf{Comprehensive benchmark results.}
\cref{fig:pareto_main} extends the comparison to six additional benchmarks: GQA, TextVQA, ScienceQA, POPE, VizWiz, and MMBench. \name consistently outperforms or matches baselines across these diverse tasks. On TextVQA and VizWiz, AdaLLaVA suffers severe degradation at 50\% budget (29.8\% and 34.3\%, respectively), whereas \name maintains 54.4\% and 55.1\% at the same FLOP level. On GQA, \name (59.8\% at 50\%) slightly trails AdaLLaVA-PruMerge (60.1\%) but surpasses AdaLLaVA (56.8\%). On ScienceQA and POPE, all three methods achieve comparable accuracy, with \name showing modest gains in the mid-range. On MMBench, AdaLLaVA and AdaLLaVA-PruMerge are competitive at full budget; \name achieves 62.0\% at 50\% compared to 63.1\% for AdaLLaVA-PruMerge and 63.3\% for AdaLLaVA. Overall, \name provides the most consistent Pareto frontier across benchmarks, with the largest advantages on VQA-style tasks (VQAv2, TextVQA, VizWiz) where adaptive token and compute control matter the most.

\subsection{Vision Token Controller Analysis}
\label{subsec:token_controller_analysis}
\begin{figure*}[t!]
    \centering
    \begin{minipage}{0.48\linewidth}
        \centering
        \includegraphics[width=1\linewidth]{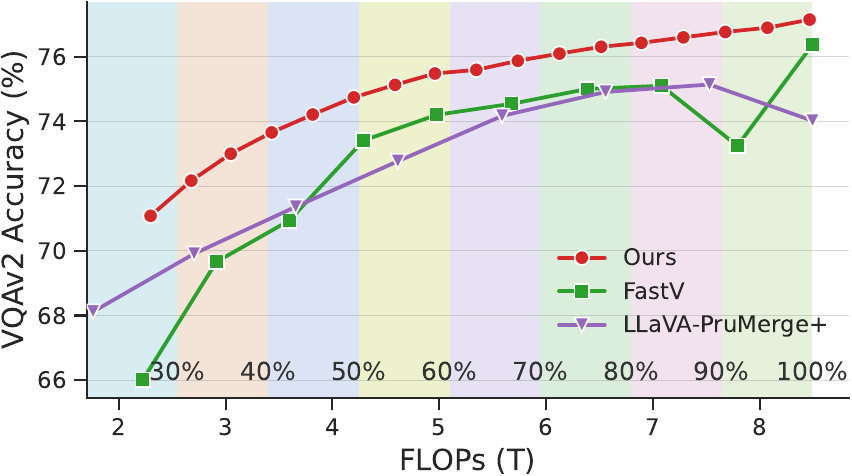}
        \caption{\textbf{Pareto frontier of token-prune methods} (VQAv2). Our learned token controller achieves a favorable tradeoff compared to FastV / LLaVA-PruMerge+ across varying budgets.}
        \label{fig:token_pareto}
    \end{minipage}
    \hfill
    \begin{minipage}{0.48\linewidth}
        \centering
        \includegraphics[width=1\linewidth]{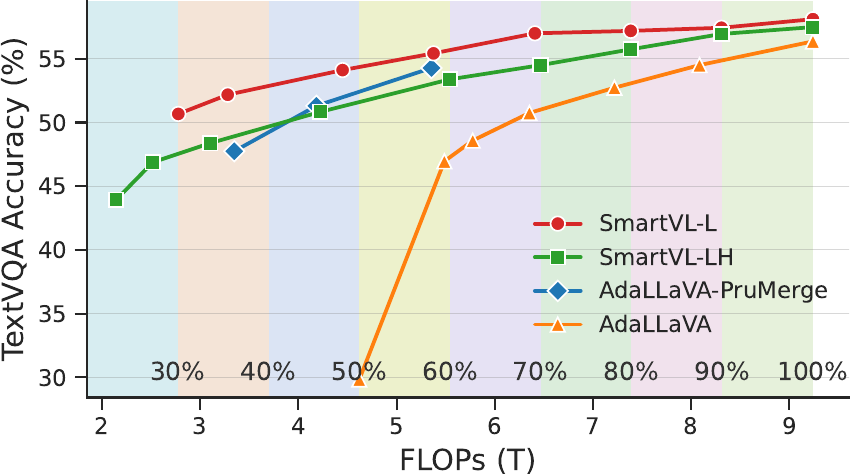}
        \caption{\textbf{Pareto frontier of different model variants} (TextVQA). Our variants (L and LH) achieve a larger adaptation range and better trade-off compared to AdaLLaVA / AdaLLaVA-PruMerge.}
        \label{fig:token_pareto2}
    \end{minipage}
\end{figure*}

A distinguishing feature of our framework relative to AdaLLaVA is the addition of a learned vision-side token controller that operates after the ViT encoder. This section isolates its contribution by comparing against established token selection methods.
\cref{fig:token_pareto} compares our token controller (with the compute controller disabled, \ie, all layers and heads active) against PruMerge+~\cite{shang2025llava} and FastV~\cite{chen2024fastv} on VQAv2. Our token controller achieves a consistently favorable accuracy-efficiency tradeoff across the full range. In the high FLOPs regime, it preserves accuracy better than both baselines by retaining the most task-relevant tokens. In the low FLOPs regime, it outperforms FastV and PruMerge+ by a clear margin. Importantly, our token controller is budget-conditioned: a single trained model produces different token subsets for different budgets and selects input-dependent token subsets for each target budget. By comparison, the fixed-pruning baselines form discrete retention-ratio operating points, while our controller changes the retained token set with the target budget. This budget-conditioned flexibility of our approach is what enables seamless integration with the compute controller in the full framework.

\subsection{Ablation Studies}
\label{subsec:ablation}

\noindent\textbf{L vs.\ LH variant.}
Comparing the two variants and baselines on TextVQA (Fig.~\ref{fig:token_pareto2}), \name yields consistent gains across all compute budgets. At 50\% FLOPs, SmartVL-L has 54.1\% accuracy, beating AdaLLaVA-PruMerge (51.3\%) and avoiding the severe degradation of AdaLLaVA (29.8\%). This performance lead persists at 70\% FLOPs (57.0\% vs. 54.3\% and 50.8\%, respectively) and scales to a baseline-leading 58.1\% at full capacity. The SmartVL-LH variant offers a viable alternative, trading a marginal drop at full budget (57.5\%). This finding indicates that MLLM redundancy primarily resides in sequence and depth. Preserving full-width attention remains essential for inference quality, making joint token-layer control outperform a fully unconstrained 3D controller. While comparable in the mid-range, AdaLLaVA-PruMerge lacks continuous adaptation, relying instead on manual, discrete configurations (\eg, 25\% tokens with 65-100\% layers). Furthermore, AdaLLaVA suffers a severe accuracy degradation at 50\% compute, exposing the fundamental limits of layer-only pruning relative to cross-dimensional control.

\begin{figure}[t!]
    \centering
    \includegraphics[width=0.95\linewidth]{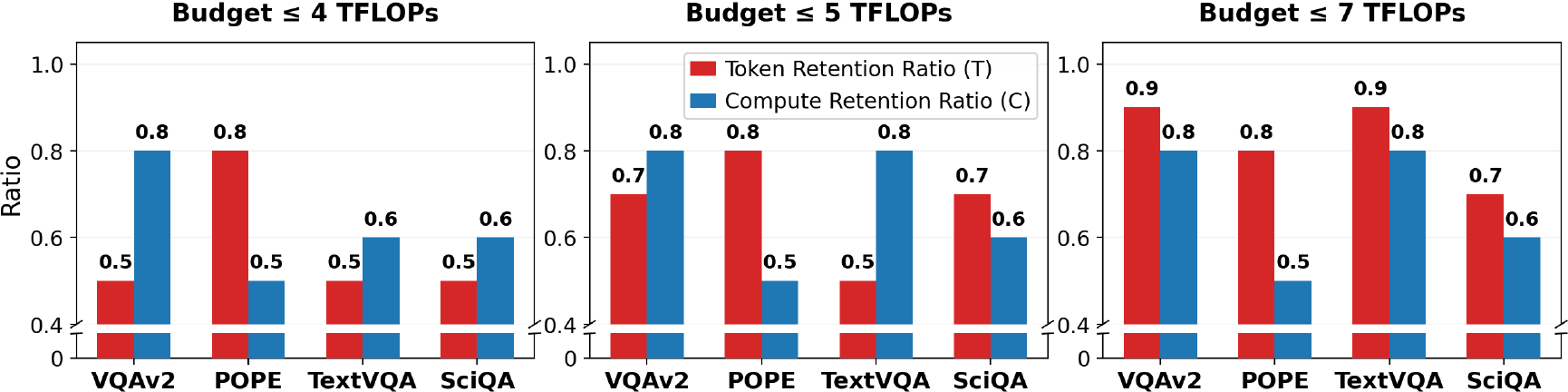}
    \caption{\textbf{Optimal token-compute adaptation varies based on task.} Best performance allocations under varying compute budgets are shown across the figures. The optimal combination differs across tasks and compute budget levels; \eg, POPE consistently favors a high token retention ratio with greater tolerance to reduced compute retention, while VQAv2 requires deeper LLM processing with a higher compute retention ratio. No single configuration or trade-off is optimal for different tasks.}
    \label{fig:teaser_benchmark_budget_comparison}
\end{figure}

\begin{figure}[t!]
    \centering
    \includegraphics[width=0.95\linewidth]{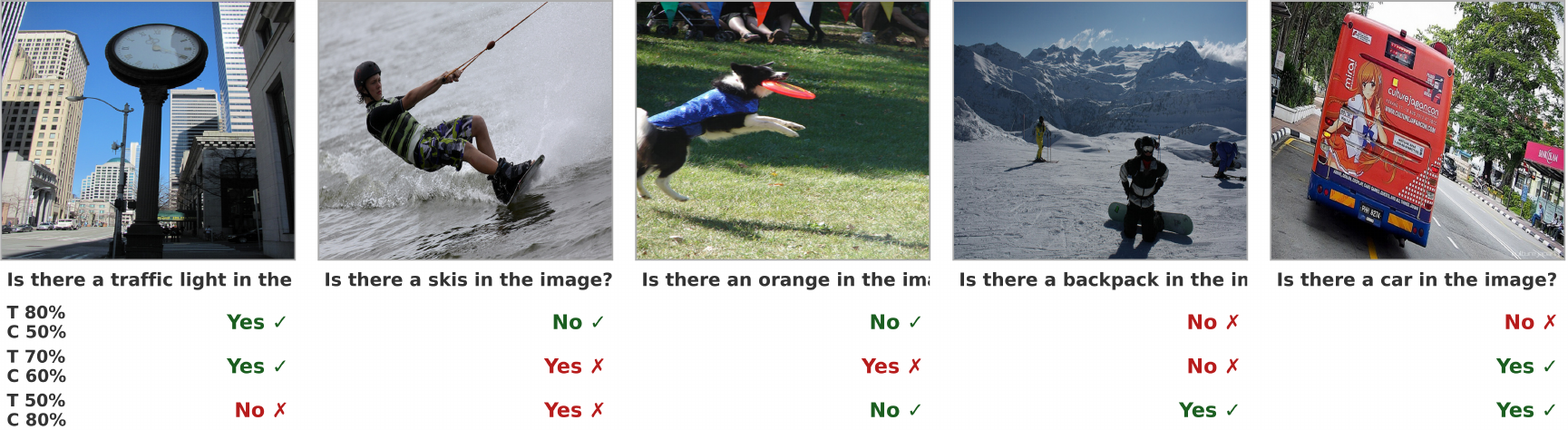}
    \includegraphics[width=0.95\linewidth]{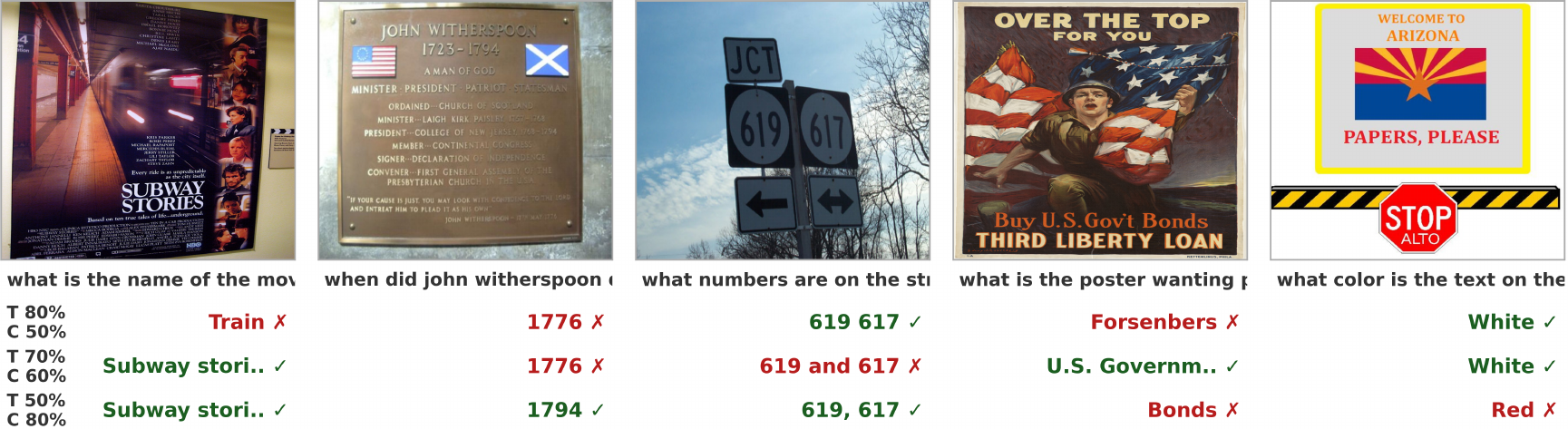}
    \caption{\textbf{Visualization of token-compute allocation under a 4 TFLOPs comptute budget}. The upper and lower figures show the prediction results on POPE and TextVQA. Different $(T, C)$ combinations exhibit varying performance depending on the input, demonstrating that the best budget allocation is content-dependent.}
    \label{fig:teaser_sample_based}
\end{figure}

\smallskip
\noindent\textbf{Content-dependent optimal token-compute allocation.}
A key design choice of \name is to jointly train the token and compute controllers, rather than optimizing either dimension in isolation. To validate this decision, we conduct an ablation study with separately trained vision-token and compute controllers. We sweep the token retention ratio $T \in \{0.5, 0.6, 0.7, 0.8, 0.9\}$ and compute retention ratio $C \in \{0.5, 0.6, 0.7, 0.8\}$, and record the resulting computation and accuracy. In this setup, each controller receives its own budget signal, specifying how many visual tokens or how much LLM compute to keep.
As shown in \cref{fig:teaser_benchmark_budget_comparison}, the optimal $(T, C)$ varies substantially across tasks and budget levels. For instance, POPE consistently favors high visual coverage with minimal LLM depth ($T{=}0.8$, $C{=}0.5$), while VQAv2 and TextVQA require progressively deeper reasoning as the budget grows ($T{=}0.9$, $C{=}0.8$ at 7 TFLOPs).
In addition, single-dimensional adaptation is not sufficient. In \cref{fig:teaser_sample_based}, we present combinations with different token or compute retention ratios but similar end-to-end FLOPs. At comparable FLOPs around 4 TFLOPs, shifting the budget from vision tokens to LLM compute (or vice versa) leads to different predictions on the same input, and the pattern also differs across tasks. This confirms that inference cost should be allocated across both dimensions in a \textit{content-aware manner}. If only one dimension were used, the model would lack the flexibility needed for diverse task types and input characteristics.

\smallskip
\noindent\textbf{Scaling to 13B models.}
We further instantiate \name on the LLaVA-1.5-13B backbone, which has 40 language layers with a fixed 20-layer prefix, and evaluate on the VQAv2 validation set. The 13B variant follows the same global-budget training and inference pipeline as the 7B model, with AdaLLaVA 13B~\cite{xu2025learning} serving as the corresponding compute-only baseline. 

\begin{wrapfigure}[12]{r}{0.50\textwidth}
    \centering
    \includegraphics[width=1\linewidth, trim=0 6mm 0 16mm]{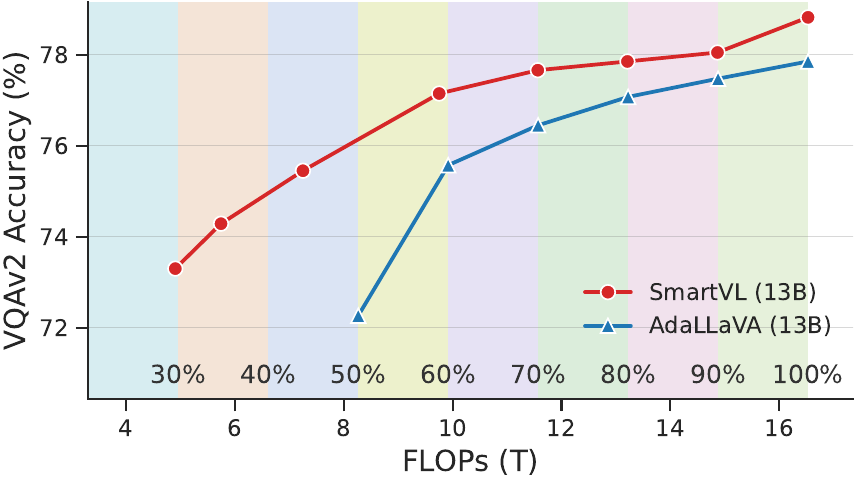}
    \caption{\small \textbf{Scaling to 13B on VQAv2.} \name improves the accuracy-efficiency Pareto frontier over AdaLLaVA.}
    \label{fig:ablation_13b}
\end{wrapfigure}

As shown in \cref{fig:ablation_13b}, joint token-compute adaptation improves the Pareto frontier at this larger scale. Near the half-budget regime, \name reaches 75.45\% accuracy at 7.25T prefill FLOPs, while AdaLLaVA reaches 72.27\% at 8.27T. At comparable mid/high compute levels, \name obtains 77.66\% at 11.56T compared with 76.45\% at 11.57T for AdaLLaVA, and reaches 78.82\% at full budget compared with 77.85\%. These results show that the learned allocation strategy transfers to a larger backbone, with joint visual-token and LLM-depth adaptation continuing to improve the accuracy--efficiency tradeoff under constrained computation.

\section{Conclusion and Discussion}
\label{sec:conclusion}

In this paper, we presented \name, a unified adaptive inference framework that jointly controls visual token usage and LLM computation in MLLMs under varying global budgets and input contents.
By coupling a vision-side token controller with an LLM-side compute controller through a shared budget encoding and a differentiable latency estimator, \name learns content-aware allocation strategies across different computational dimensions.
Extensive experiments across multiple benchmarks demonstrated that this joint scheduling achieves stronger accuracy-efficiency tradeoffs than prior adaptive baselines.

\smallskip
\noindent\textbf{Future work.}
Our latency estimator targets prefill-stage FLOPs as the primary inference cost; extending this adaptive routing to the memory-bound decode stage remains open. Our architectural routing is orthogonal to weight quantization, and combining the two may yield compounded gains. Finally, translating FLOP reductions into wall-clock improvements will require integrating sparse execution masks with specialized hardware kernels.

\medskip
\noindent\textbf{Acknowledgements.}
This material is based in part upon work supported by the National Science Foundation under Grant Numbers CNS-2333487/2333491 (CPS Frontier), CNS-2146449 (NSF CAREER), and IIS-2442739 (NSF CAREER), as well as a gift from Google. Any opinions, findings, conclusions, or recommendations expressed in this material are those of the authors and do not necessarily reflect the views of the sponsors. We thank the reviewers and area chairs for their constructive feedback, which helped improve this work.

\bibliographystyle{splncs04}
\bibliography{6-references.bib}

\end{document}